\documentclass[runningheads]{llncs}
\usepackage[table,xcdraw]{xcolor}
\usepackage{todonotes}
\usepackage{amsmath}
\usepackage{graphicx}
\usepackage{comment}
\usepackage{caption}
\usepackage{subcaption}
\usepackage{multirow}
\usepackage{cite}
\usepackage{booktabs}

\usepackage{array}
\usepackage{amssymb}
\usepackage{placeins}
\usepackage{soul}
\usepackage{tabularx, makecell}
\usepackage{pifont}

\begin{document}
%


\title{Test-Time Selection\\for Robust Skin Lesion Analysis}

%
%
\author{Alceu Bissoto\inst{1} \and
Catarina Barata\inst{2} \and
Eduardo Valle\inst{3,4} \and Sandra Avila\inst{1}}
\authorrunning{Bissoto et al.}

\institute{Recod.ai Lab, Institute of Computing, University of Campinas, Brazil
\and Institute for Systems and Robotics, Instituto Superior Técnico, Portugal \and School of Electrical and Computing Engineering, University of Campinas, Brazil \and Valeo.ai Paris}
%
%

\maketitle              
\begin{abstract}
Skin lesion analysis models are biased by artifacts placed during image acquisition, which influence model predictions despite carrying no clinical information. Solutions that address this problem by regularizing models to prevent learning those spurious features achieve only partial success, and existing test-time debiasing techniques are inappropriate for skin lesion analysis due to either making unrealistic assumptions on the distribution of test data or requiring laborious  annotation from medical practitioners.  
We propose TTS (Test-Time Selection), a human-in-the-loop method that leverages positive (e.g., lesion area) and negative (e.g., artifacts) keypoints in test samples. 
TTS effectively steers models away from exploiting spurious artifact-related correlations without retraining, and with less annotation requirements. Our solution is robust to a varying availability of annotations, 
and different levels of bias. We showcase on the ISIC2019 dataset (for which we release a subset of annotated images) how our model could be deployed in the real-world for mitigating bias. 

\end{abstract}


\section{Introduction}
Spurious correlations between conspicuous image features and annotation labels are easy to learn, but since they have no actual predictive power they compromise the robustness of models. In medical image analysis, with datasets much smaller than the typical computer vision state-of-the-art, their effect is increased.  
In skin lesion analysis, one of the most studied confounders are artifacts produced during image acquisition (such as rulers, color patches, and ink markings). Even if the correlation of the presence of each artifact with the lesion diagnostic is small, the combined effect suffices to distract models from clinically-robust features \cite{bissoto2019constructing, bissoto2020debiasing, combalia2022validation}. 
Mitigating bias during training is an active research area, but methods still struggle to surpass strong baselines~\cite{gulrajani2020search}. A complementary solution is to change the inference procedure to mitigate biases during test~\cite{bissoto2023even}. For that, solutions have exploited test batch statistics for feature alignment~\cite{wang2021tent, iwasawa2021t3a}. However, test batch statistics heavily rely on the batch size (the bigger, the better) and on the homogeneity of the test distribution. For medical data, one attractive option is to exploit (few or quickly obtainable) extra annotations to infuse domain knowledge into the models' predictions, increasing model robustness and trust of medical practitioners~\cite{daneshjou2022checklist}. 

In comparison to other medical fields, skin lesion analysis researchers have access to rich annotations to support this path. Besides high-quality images, there are available annotations regarding segmentation masks, dermoscopic attributes, the presence of artifacts, and other clinical information such as age, sex, and lesions' anatomical site. In particular, segmentation masks experience the most success, granting more robustness to classification. We build upon this success to create a solution that dependd on human-defined keypoints, which are far cheaper to annotate than lesion segmentation masks.

\begin{figure}[t]
    \centering
    \includegraphics[trim={0 4cm 0 3cm},clip, width=0.85\textwidth]{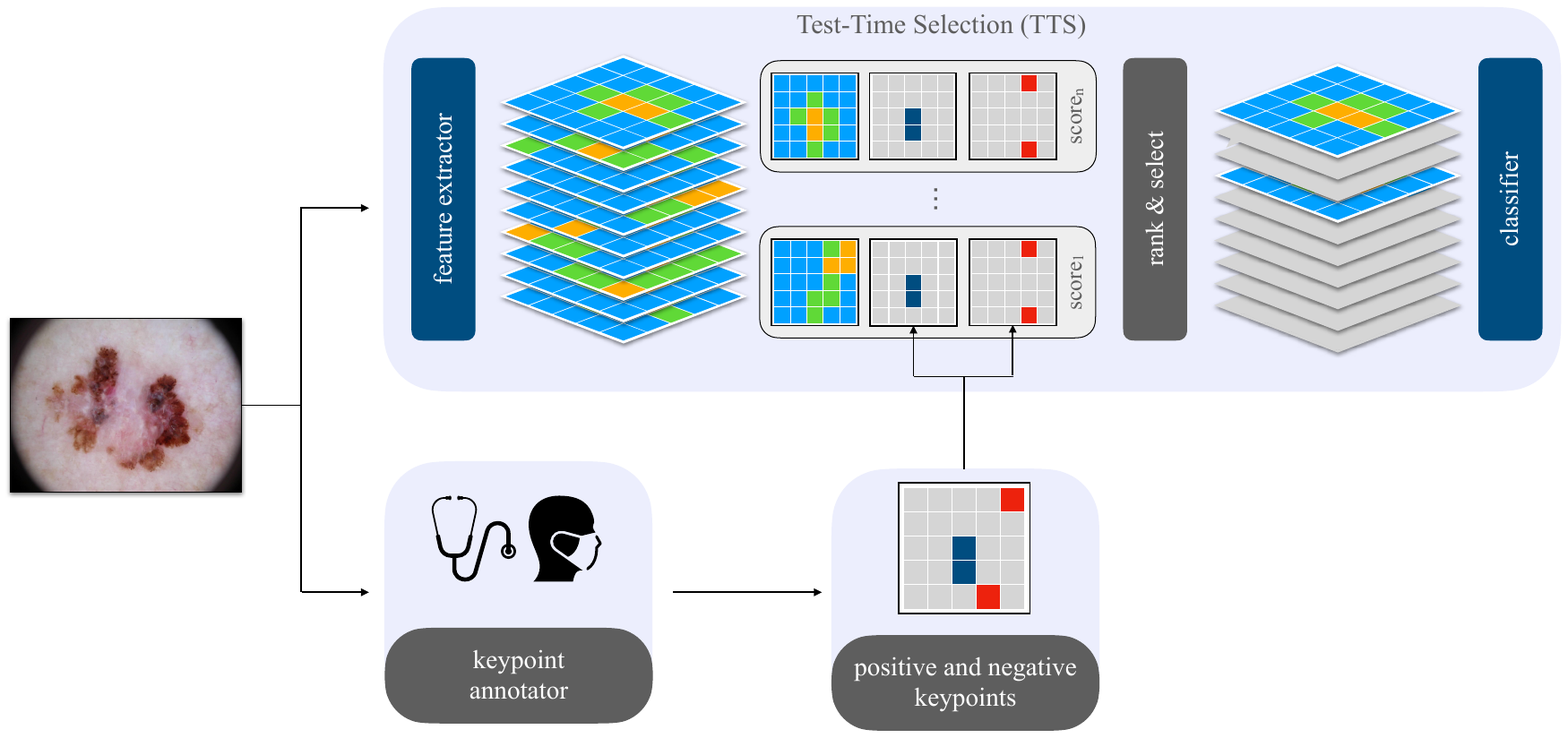}
    \caption{Test-Time Selection (TTS). An annotator provides negative (background, artifacts) and positive (lesion area) keypoints, used to rank and select activation units from the last layer of the pretrained feature extractor. Features related to negative keypoints are masked to zero.}
    \label{fig:summary}
\end{figure}

In this work, we propose TTS (Test-Time Selection), a method to incorporate human-defined points of interest into trained models to select robust features during test-time. In Fig. \ref{fig:summary}, we show a summary of our method. In more detail, we first gather human-selected keypoints of interest (positive and negative). Then, we rank the last layer activation units based on their affinity to the keypoints. Finally, we mute (set to zero) the 90\% worst features, using only the remaining 10\% for classification. There are no changes to the models' weights, making this procedure lightweight and easy to integrate in different pipelines.

Our method is compatible with the daily clinical routine to avoid overwhelming medical practitioners with the technology that is intended to assist them. The human intervention must be as quick and straightforward as possible while granting enough information to steer models away from spurious correlations. We show that we can improve robustness even from a single pair of positive and negative interest points that merely identify lesion and background, and achieve stronger results by using the location of artifacts.

We summarize our contributions as follows:\vspace{-0.1cm}
\begin{itemize}
    \item We propose a method for test-time selection based on human-defined criteria that boosts the robustness of skin lesion analysis models\footnote{Code is available at \url{https://github.com/alceubissoto/skin-tts}.}.
    \item We show that our method is effective throughout different bias levels.
    \item We show that a single positive and negative interest point is sufficient to improve significantly models' robustness.
    \item We manually annotate the position of artifacts in skin lesion images and use these selected keypoints in our solution, further improving performance.
\end{itemize}

\section{Related Work}

Test-time debiasing can adapt deep learning to specific population characteristics and hospital procedures that differ from the original dataset. 
Most methods for test-time debiasing exploit statistics of the batch of test examples. Tent~\cite{wang2021tent} (Test entropy minimization) proposes to update batch normalization weights and biases to minimize the entropy of the test batch. Similarly, T3A~\cite{iwasawa2021t3a} (Test-Time Template Adjuster) maintains new class prototypes for the classification problem, which are updated with test samples, and finally used for grounding new predictions. Both approaches rely on two strong assumptions: that a large test batch is available during prediction, and that all test samples originate from the same distribution.

Those assumptions fail for medical scenarios, where diagnostics may be performed one by one, and populations attending a given center may be highly multimodal. To attempt to work in this more challenging scenario, SAR~\cite{niu2023sar} (Sharpness-Aware and Reliable optimization scheme) proposes to perform entropy minimization updates considering samples that provide more stable gradients while finding a flat minimum that brings robustness regarding the noisy samples. Despite showing good performances in corrupted scenarios (e.g., ImageNet-C~\cite{hendrycks2019robustness}), SAR is heavily dependent on the model's architecture, being inappropriate for models with batch normalization layers. 
In contrast with these methods, our solution does not use any test batch statistic, does not require training nor updates to the models' weights, and does not rely upon any particular architecture structure to improve performance.

Another approach is to change the network's inputs to remove biasing factors. NoiseCrop~\cite{bissoto2022} showed considerable robustness improvements for skin lesion analysis by using skin segmentation masks to replace the inputs' backgrounds with a common Gaussian noise. Despite its benefits, NoiseCrop is hard to integrate into clinical practice as it depends on laborious segmentation masks annotated by dermatologists. 
Also, NoiseCrop discards relevant information in the patient's healthy skin and introduces visual patterns that create a distribution shift of its own.
Our solution does not suffer from these problems since our intervention takes place in feature space, and we show it is effective using very few keypoints. We summarize the differences between our method and the literature in Table~\ref{tab:comparison}.

\section{Methodology}

Previous works showed the potential of test-time debiasing, but depended on weight updates using test batches statistics~\cite{wang2021tent, iwasawa2021t3a} and architecture components~\cite{niu2023sar}. We decided instead to use human feedback over positive and negative image keypoints to steer the models. We aimed at making the annotation procedure as effortless as possible, allowing to integrate the method into the clinical practice of skin lesion analysis. The resulting Test-Time Selection (TTS) is summarized in Fig.~\ref{fig:summary}.

\newcommand{\cmark}{\color{teal}\ding{51}}%
\newcommand{\xmark}{\color{red}\ding{55}}%

\begin{table}[t]
\centering
\caption{Comparison of TTS with state-of-the-art test-time debiasing.}
\label{tab:comparison}
\resizebox{\textwidth}{!}{
\begin{tabular}{lccccc}
\toprule
 & Tent~\cite{wang2021tent} & T3A~\cite{iwasawa2021t3a} & SAR~\cite{niu2023sar} & NoiseCrop~\cite{bissoto2022} & TTS (Ours)\\ 
\midrule
Test-time only & \cmark & \cmark & \cmark & \cmark & \cmark \\
Human-in-the loop & \xmark & \xmark & \xmark & \cmark & \cmark \\
No parameter updates & \xmark & \xmark & \xmark & \cmark & \cmark \\
Robust to changing test statistics & \xmark & \xmark & \cmark & \cmark & \cmark \\
Architecture agnostic & \xmark & \cmark & \xmark & \cmark & \cmark \\
Few extra information required & -- & -- & -- & \xmark & \cmark \\ 
\bottomrule
\end{tabular}%
}
\end{table}

\noindent{\textbf{\textit{TTS: Test-Time feature Selection.}}}
We assume access to a single test sample $x$, associated with a set of positive  $K_p = \{kp_1, kp_2, ..., kp_p\}$ and negative $K_n = \{kn_1, kn_2, ..., kn_n\}$ human-selected keypoints on the image. The positive keypoints represents areas of the image that should receive more attention (e.g., the lesion area), while the negative points represent area of the image that should be ignored (e.g., the background, or spurious artifacts). We denote the feature extractor from a pretrained neural network by $f(\cdot)$, and the associated classifier by $g(\cdot)$. 

For each image $x$, the feature extractor generates a representation $f(x)$, which is upsampled to match the original image $x$ size for test-time selection. 
For each channel $c$ in $f(x)$, we extract the values corresponding to the coordinates specified by the keypoints and compute their sums $S_{p_c} = \sum_{k \in K_p} f(x)_{c}[k]$, and $S_{n_c} = \sum_{k \in K_n} f(x)_{c}[k]$,
where $f(x)_{c}[k]$ denotes the value at the keypoint $k$ for channel $c$ of $f(x)$.
We calculate a score $S_c$ for each channel $c$ as the difference between the sums of the representations at the positive and negative keypoints:\vspace{-0.1cm}
\begin{equation}
S_c = \alpha S_{p_c} - (1 - \alpha) S_{n_c},
\end{equation}
\noindent where $\alpha$ controls the strength of the positive and negative factors. We use $\alpha=0.4$ to give slightly more weight to the negative keypoints related to the sources of bias (i.e., artifacts) investigated in this work. If the keypoint annotation confidently locates positive or negative points of interest, $\alpha$ can be adjusted to give it more~weight.

We use the scores to rank the channels with higher affinity to the input keypoints. We define a set $T$ which consists of the indices corresponding to the top $\lambda\%$ scores in $S_c$, i.e., $T = \{c : S_c \text{ is among the top } \lambda\% \text{ of scores}\}$. In other words, $\lambda$ controls how much information is muted. In general, we want to mute as much as possible to avoid using spurious correlations.
In our setup, we keep only 10\% of the original activation units. 
Next, we form a binary mask $M$ with values $m_c$ defined as: $m_c = 1, \text{if } c \in T$, or $m_c = 0, \text{if } c \notin T$.

Finally, the masked version of $f(x)$, denoted as $f'(x)$, is computed by $f(x)$: $f'(x) = f(x) \odot M$, where $\odot$ represents the element-wise multiplication.
The masked feature map $f'(x)$ is the input for our neural network's classifier component $g(\cdot)$, which yields the final prediction. As such procedure happens individually for each dataset sample, different samples can use the activation units that best suit it, which we verified to be crucial for the effectiveness of this~method.

\noindent{\textbf{\textit{Keypoints.}}}
We always assume having access to the same number of positive and negative keypoints (i.e., for 2 keypoints, we have one positive and one negative). 
We explore two options when selecting keypoints. The first option is more general and adaptable for most computer vision problems: Positive keypoints represent the foreground target object (e.g., lesion), while negative keypoints are placed in the background. To extract these keypoints we make use of skin lesion segmentation masks\footnote{We employ the ground-truth segmentation masks when available, and infer the segmentation with a deep learning model~\cite{chen2018encoder} when they are not.}. Using keypoints instead of the whole mask lessens the impact of mask disagreement (from annotators or segmentation models) in the final solution.

The second option uses domain knowledge to steer the model's prediction further: instead of sampling negative keypoints from the background, we restrict the points to the artifacts. The main benefit is allowing models to consider the skin areas around the lesion, which can provide clinically meaningful features. For that option, we manually annotate the samples on our test sets, adding negative keypoints on 4 types of artifacts: dark corners, rulers, ink markings, and patches. Other types artifacts (hair, gel bubbles, and gel borders) are hard to describe with few keypoints, and were not keypoint-annotated, but were used for trap set separation. This fine-grained annotation, allows us to boost the importance of negative keypoints by setting $\alpha$ to $0.2$, for example.

\noindent{\textbf{\textit{Data and experimental setup.}}} 
We employ the ISIC 2019~\cite{isic2019data, codella2018skin, tschandl2018ham10000} dataset. The class labels are selected and grouped such that the task is always a binary classification of melanoma \textit{vs.}~benign (other, except for carcinomas). We removed from all analysis samples labeled basal cell carcinoma or squamous cell carcinoma.
Train and test sets follow the ``trap set'' separation introduced by Bissoto et al.~\cite{bissoto2020debiasing, bissoto2022}, that craft training and test sets where the correlations are amplified between artifacts and the malignancy of the skin lesion, at the same time that correlations in train and test are opposite. 
Trap sets follow a factor that controls the level of bias, from 0 (randomly selected sets) to 1 (highly biased). In detail, for each sample, the factor controls the probability of following the train-test separation suggested by the solver or assigning it randomly to~an~environment. 

All our models consider Empirical Risk Minimization~\cite{vapnik1992principles} as the training objective. Our baseline is doing test-time augmentation with 50 replicas, a standard in skin lesion analysis~\cite{perez2018data}. For a more realistic clinical setup, we always assume to have access to a single test image at each time. The results for TTS also perform test-time augmentation with 50 replicas, showing that our model can easily be combined with other test-time inference techniques. The pretrained models used for all the experiments were fine-tuned for 100 epochs with SGD with momentum, selecting the checkpoint based on validation performance\footnote{For choosing these models hyperparameters, we performed a grid-search over learning rate (values 0.00001, 0.0001, 0.001), and weight-decay (0.001, 0.01, 0.1, 1.0), for 2 runs on a validation set randomly split from the training set.}. Conventional data augmentation (e.g., vertical and horizontal shifts, color jitter) are used as training and testing. All results refer to the average of 5 runs (each with a different training/validation/test partition\footnote{Training/validation/test contains 60/10/30\% of the total data.} and random seed). 

\section{Results}

\begin{figure}[bt]
    \centering
    \includegraphics[width=.775\textwidth]{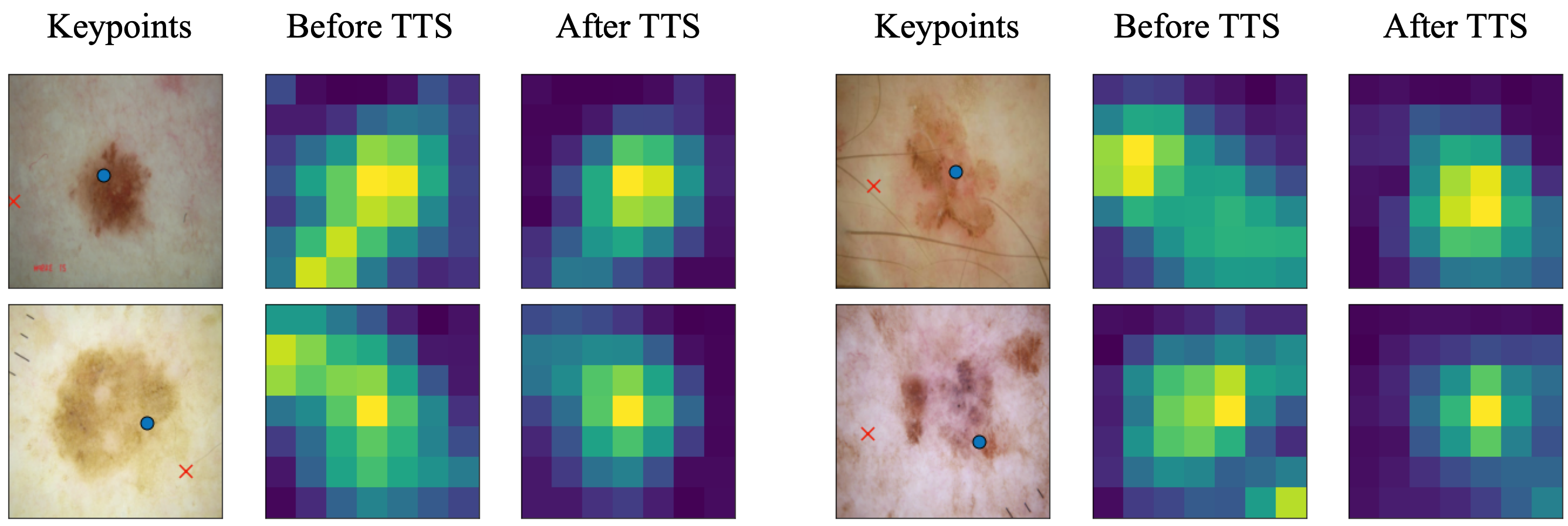}
    \caption{Attention maps before and after our feature selection. Using a few keypoint annotations, TTS successfully reduces the importance of spurious features in the background, shifting the model's focus to the lesion.}
    \label{fig:vis}
\end{figure}

We show our main results in Table \ref{tab:main_results}, comparing our solution with the state-of-the-art of test-time adaptation. All models are evaluated in trap sets, which create increasingly hard train and test partitions. On training, biases are amplified. On test, the correlations between artifacts and labels are shifted, punishing the models for using the biases amplified on training. The ``training bias'' controls the difficulty, being 1.0 as the hardest case. In this scenario, traditional trained models, even with test-time augmentation, abdicate from learning robust features and rely entirely on biases. Despite NoiseCrop~\cite{bissoto2022} can highly improve the performance, it requires the whole segmentation mask, which is expensive to annotate and suffer from low inter-annotator agreement issues~\cite{ribeiro2020less}. We show that TTS consistently surpasses baselines using very few annotated keypoints. By analyzing the attention maps before and after our procedure (Fig.~\ref{fig:vis}), TTS successfully mitigates bias, diminishing the importance of artifacts.
Also, its flexibility allows for better results once the annotated keypoints locate the artifacts biasing the solution (e.g., dark corners, rulers, ink markings, and~patches).

\begin{table}[t]
\caption{Main results and ablations (on number and annotation source of keypoints) for the hardest trap tests (training bias = 1.0). TTS achieves state-of-the-art performances while using very few annotated keypoints.}%
\centering
\scriptsize
\begin{tabular}{llllll}
\toprule
& Method & \#Keypoints\hspace{0.2cm} & Annotation\hspace{0.2cm} & Alpha\hspace{0.2cm} & AUC \\
\midrule
baseline & Test-Time Aug\hspace{0.2cm} & -- & -- & -- & $58.4$ \scriptsize{$\pm 1.6$}\\
literature\hspace{0.2cm} & T3A~\cite{iwasawa2021t3a} & -- & -- & -- & $56.7$ \scriptsize{$\pm 3.2$}\\
literature & Tent~\cite{wang2021tent} & -- & -- & -- & $54.1$ \scriptsize{$\pm 14.5 $}\\
literature & NoiseCrop~\cite{bissoto2022} & 50,176 & segm. mask & -- & $72.7$ \scriptsize{$\pm 3.1$}\\
\midrule
& TTS (ours)  & 40 & artifacts & 0.2 & $\textbf{75.0}$\hspace{0.04cm}\scriptsize{$\pm 1.1$}\\
ablation & TTS (ours) & 2 & segm. mask & 0.4 & $68.2$ \scriptsize{$\pm 1.5$}\\

ablation & & 10 & segm. mask & 0.4 & $71.6$ \scriptsize{$\pm 2.2$}\\

ablation & & 20 & segm. mask & 0.4 & $72.9$ \scriptsize{$\pm 2.4$}\\

ablation & & 40 & segm. mask & 0.4 & $73.3$ \scriptsize{$\pm 2.6$}\\

ablation & & 100 & segm. mask & 0.4 & $73.9$ \scriptsize{$\pm 2.5$}\\
ablation & & 2 & artifacts & 0.4 & $69.6$ \scriptsize{$\pm 1.1$}\\

ablation & & 40 & artifacts & 0.4 & $73.3$ \scriptsize{$\pm 0.9$}\\

ablation & & 2 & artifacts & 0.2 & $72.2$ \scriptsize{$\pm 0.9$}\\
\bottomrule
\end{tabular}
\label{tab:main_results}
\end{table}

\noindent{\textbf{\textit{Amount of available keypoints.}}}
We evaluate the effect of limiting the availability of keypoints. This is an essential experiment for assessing the method's clinical applicability. If it requires too many points to be effective, it may overwhelm clinical practitioners with annotating duties, which beats the purpose of using computer-assisted diagnosis systems. In Table \ref{tab:main_results}, we show that our method can positively impact the robustness of pretrained models even in extreme conditions where a single negative and positive keypoint is annotated. Aside from the minimum impact in the clinical pipeline, it also shows to be robust to different annotators since the improvements are consistent by sampling positive and negative keypoints at random from segmentation~masks. 

\noindent{\textbf{\textit{Keypoint annotation granularity.}}}
The flexibility of using keypoints (instead of full segmentation masks) not only allows for easy inclusion in the daily clinical routine, but also allows for fine-grained concepts to be annotated without being time-consuming. In this experiment, we manually annotated the trap test sets with keypoints that locate $4$ artifacts: dark corner, ruler, ink markings, and patches. With fine-grained annotations of artifacts to provide negative keypoints, we can increase negative keypoints importance by shrinking $\alpha$, achieving our best result. We show our results in Table \ref{tab:main_results}.

Using artifact-specific keypoints instead of background ones does not punish models for using the lesions' surrounding skin in the decision process, being beneficial for diagnosis classes such as actinic keratosis, where the skin itself provide clinically-meaningful information. This change further boosts previous gains both when 1 or 20 positive and negative points were available. Our method can be used in other scenarios, where not only negative but relevant positive information can be encouraged to be used by models, such as the presence of dermoscopic attributes.

\begin{figure}[t]
    \centering
    \begin{minipage}{0.45\textwidth}
        \centering
        \includegraphics[trim={1cm 0 1cm 1cm},clip,width=1.1\textwidth]{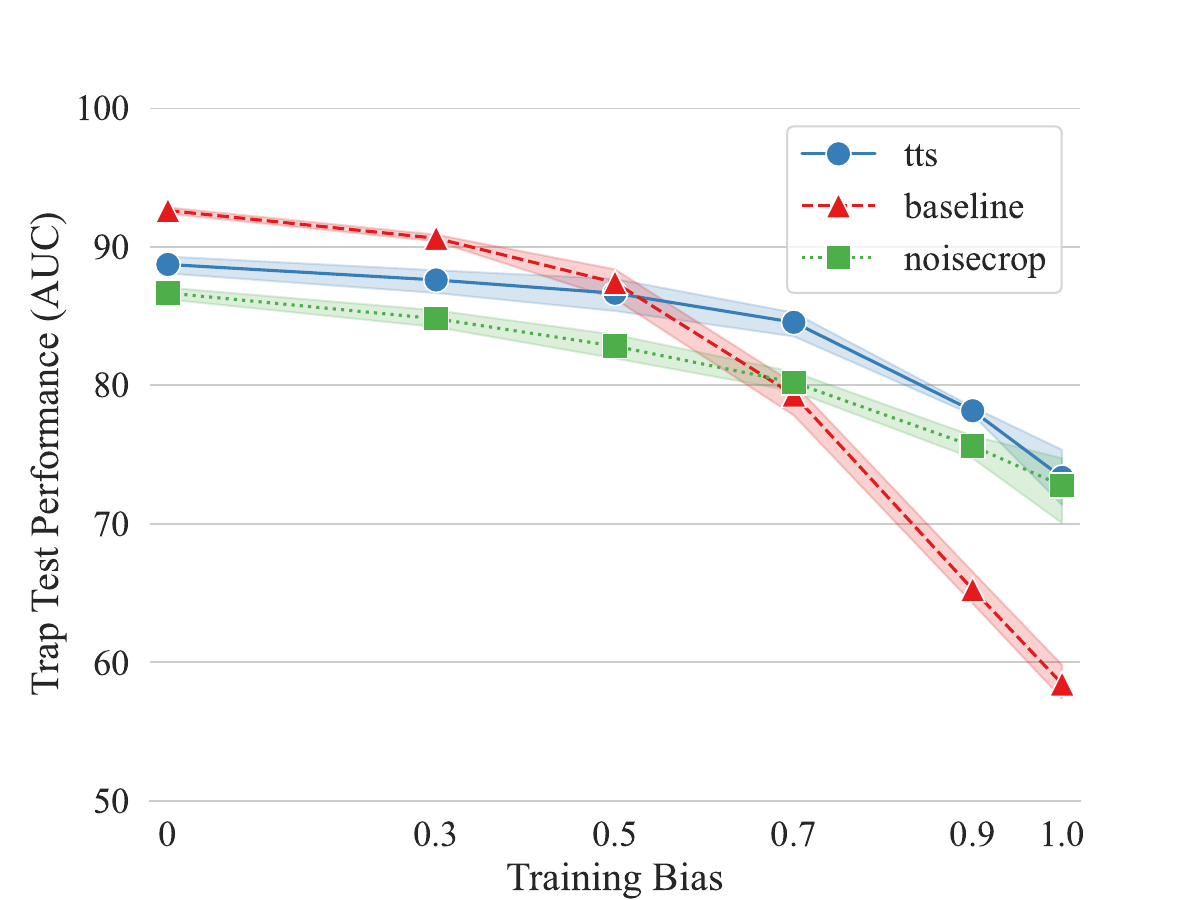}
    \end{minipage}\hfill
    \begin{minipage}{0.45\textwidth}
        \caption{Ablation of our TTS over different intensities of bias. TTS consistently outperforms NoiseCrop~\cite{bissoto2022} across bias intensities while using a fraction of the extra-information available: NoiseCrop uses the whole segmentation mask, while in this example, we use 20 positive and 20~negative keypoints.}
        \label{fig:bfs}
    \end{minipage}
\end{figure}

\noindent{\textbf{\textit{Different levels of bias.}}}
We evaluate our solution over different levels of bias from trap sets. 
Trap sets allow a better assessment of models ability to generalize. As the training bias increases, the task becomes increasingly hard for the model, as correlations between artifacts and labels get harder to pass unnoticed. At the same time, the higher the bias factor, the better trap test does at punishing the model for exploiting spurious correlations. When the training bias is low, robust models are expected to achieve a worse result than unbounded ones, as exploiting spurious correlations is rewarded by evaluation metrics. However, even if we can not perfectly measure the bias reliance in intermediate bias, performing well in these situations is crucial since real-world scenarios might not present exaggerated biases. In Fig. \ref{fig:bfs}, we show that our solution outperforms NoiseCrop across all bias factors. We hypothesize that NoiseCrop introduces a distribution shift when it replaces the inputs' background with noise. We avoid this shortcoming by intervening in the feature space instead of the pixel space, which proved robust to the sparsity induced by our procedure.

\section{Conclusion}

We propose a method for test-time debiasing of skin lesion analysis models, dealing with biases created by the presence of artifacts on the ISIC 2019 dataset. 
Our method select features during inference taking user-defined keypoints as a guide to mute activation units. We show that our method encourages the attention map focus more on lesions, translating to higher performance on biased scenarios. We show that our model is effective throughout different levels of bias even with single pair of annotated keypoints, thus allowing frugal human-in-the-loop learning. It benefits from fine-grained annotations, such as artifact locations, and is lightweight as it does not require training.
In future works, we want to explore the possibility of keeping a memory bank of important previously annotated concepts to consider before each prediction.  Muting features is a general principle, extensible to other data modalities, including text (e.g., from medical summaries), an idea that we would also like to explore in the future.

\vspace{0.25cm}
\noindent \textbf{Acknowledgments}: 
A. Bissoto is funded by FAPESP (2019/19619-7, 2022/ 09606-8). 
C. Barata is funded by the FCT projects LARSyS (UID/50009/2020), CEECIND/00326/2017, and Center for Responsible AI C645008882-00000055.
S.~Avila is funded by CNPq 315231/2020-3, FAPESP 2013/08293-7, 2020/09838-0, H.IAAC, Google LARA 2021 and AIR~2022.

%
%
%
\bibliographystyle{splncs04}
\bibliography{ref}

\begin{thebibliography}{10}
\providecommand{\url}[1]{\texttt{#1}}
\providecommand{\urlprefix}{URL }
\providecommand{\doi}[1]{https://doi.org/#1}

\bibitem{bissoto2019constructing}
Bissoto, A., Fornaciali, M., Valle, E., Avila, S.: ({D}e){C}onstructing bias on
  skin lesion datasets. In: IEEE Conference on Computer Vision and Pattern
  Recognition Workshops (CVPRW) (2019)

\bibitem{bissoto2020debiasing}
Bissoto, A., Valle, E., Avila, S.: Debiasing skin lesion datasets and models?
  not so fast. In: IEEE Conference on Computer Vision and Pattern Recognition
  Workshops (CVPRW) (2020)

\bibitem{bissoto2022}
Bissoto, A., Barata, C., Valle, E., Avila, S.: Artifact-based domain
  generalization of skin lesion models. In: European Conference on Computer
  Vision Workshops (ECCVW) (2023)

\bibitem{bissoto2023even}
Bissoto, A., Barata, C., Valle, E., Avila, S.: Even small correlation and
  diversity shifts pose dataset-bias issues. arXiv preprint arXiv:2305.05807
  (2023)

\bibitem{chen2018encoder}
Chen, L.C., Zhu, Y., Papandreou, G., Schroff, F., Adam, H.: Encoder-decoder
  with atrous separable convolution for semantic image segmentation. In:
  European Conference on Computer Vision (ECCV) (2018)

\bibitem{codella2018skin}
Codella, N.C., Gutman, D., Celebi, M.E., Helba, B., Marchetti, M.A., Dusza,
  S.W., Kalloo, A., Liopyris, K., Mishra, N., Kittler, H., et~al.: Skin lesion
  analysis toward melanoma detection: A challenge at the 2017 international
  symposium on biomedical imaging (isbi), hosted by the international skin
  imaging collaboration (isic). In: IEEE International Symposium on Biomedical
  Imaging (ISBI). pp. 168--172 (2018)

\bibitem{isic2019data}
Combalia, M., Codella, N., Rotemberg, V., Helba, B., Vilaplana, V., Reiter, O.,
  Carrera, C., Barreiro, A., Halpern, A., Puig, S., et~al.: Bcn20000:
  Dermoscopic lesions in the wild. arXiv:1908.02288  (2019)

\bibitem{combalia2022validation}
Combalia, M., Codella, N., Rotemberg, V., Carrera, C., Dusza, S., Gutman, D.,
  Helba, B., Kittler, H., Kurtansky, N.R., Liopyris, K., et~al.: Validation of
  artificial intelligence prediction models for skin cancer diagnosis using
  dermoscopy images: the 2019 international skin imaging collaboration grand
  challenge. The Lancet Digital Health  \textbf{4}(5) (2022)

\bibitem{daneshjou2022checklist}
Daneshjou, R., Barata, C., Betz-Stablein, B., Celebi, M.E., Codella, N.,
  et~al.: Checklist for evaluation of image-based artificial intelligence
  reports in dermatology: Clear derm consensus guidelines from the
  international skin imaging collaboration artificial intelligence working
  group. JAMA dermatology  \textbf{158}(1) (2022)

\bibitem{gulrajani2020search}
Gulrajani, I., Lopez-Paz, D.: In search of lost domain generalization. In:
  International Conference on Learning Representations (ICLR) (2021)

\bibitem{hendrycks2019robustness}
Hendrycks, D., Dietterich, T.: Benchmarking neural network robustness to common
  corruptions and perturbations. International Conference on Learning
  Representations (ICLR)  (2019)

\bibitem{iwasawa2021t3a}
Iwasawa, Y., Matsuo, Y.: Test-time classifier adjustment module for
  model-agnostic domain generalization. Advances in Neural Information
  Processing Systems (NeurIPS)  (2021)

\bibitem{niu2023sar}
Niu, S., Wu, J., Zhang, Y., Wen, Z., Chen, Y., Zhao, P., Tan, M.: Towards
  stable test-time adaptation in dynamic wild world. In: Internetional
  Conference on Learning Representations (ICLR) (2023)

\bibitem{perez2018data}
Perez, F., Vasconcelos, C., Avila, S., Valle, E.: Data augmentation for skin
  lesion analysis. OR 2.0 Context-Aware Operating Theaters, Computer Assisted
  Robotic Endoscopy, Clinical Image-Based Procedures, and Skin Image Analysis
  (2018)

\bibitem{ribeiro2020less}
Ribeiro, V., Avila, S., Valle, E.: Less is more: Sample selection and label
  conditioning improve skin lesion segmentation. In: IEEE Conference on
  Computer Vision and Pattern Recognition Workshops (CVPRW) (2020)

\bibitem{tschandl2018ham10000}
Tschandl, P., Rosendahl, C., Kittler, H.: The ham10000 dataset, a large
  collection of multi-source dermatoscopic images of common pigmented skin
  lesions. Scientific data  \textbf{5}(1) (2018)

\bibitem{vapnik1992principles}
Vapnik, V.: Principles of risk minimization for learning theory. In: Advances
  in Neural Information Processing Systems (NeurIPS) (1992)

\bibitem{wang2021tent}
Wang, D., Shelhamer, E., Liu, S., Olshausen, B., Darrell, T.: Tent: Fully
  test-time adaptation by entropy minimization. In: International Conference on
  Learning Representations (ICLR) (2021)

\end{thebibliography}

\end{document}